\documentclass{article}
\pdfoutput=1
\usepackage{spconf}
\usepackage[utf8]{inputenc}
\usepackage{tikz}
\usepackage{pgfplots}
\usepackage{graphicx}
\usepackage{subcaption}
\usepackage{adjustbox}
\usepackage{amsmath}
\usepackage{amsthm}
\usepackage{amsfonts}
\usepackage{xcolor}
\usepackage{url}
\usepackage{multirow}

\newtheorem{definition}{Definition}

\pgfplotsset{compat=1.16}
\usetikzlibrary{pgfplots.colorbrewer,}
\pgfplotsset{
    cycle list/.define={my marks}{
        every mark/.append style={solid,fill=\pgfkeysvalueof{/pgfplots/mark list fill}},mark=*\\
        every mark/.append style={solid,fill=\pgfkeysvalueof{/pgfplots/mark list fill}},mark=square*\\
        every mark/.append style={solid,fill=\pgfkeysvalueof{/pgfplots/mark list fill}},mark=triangle*\\
        every mark/.append style={solid,fill=\pgfkeysvalueof{/pgfplots/mark list fill}},mark=diamond*\\
    },
}
\pgfplotsset{
  errorBars/.style={
    error bars/error bar style={very thick},
    error bars/error mark options={very thick,solid,mark size=3pt,rotate=90},
    error bars/y dir=both,
    error bars/y explicit,
  }
}
\pgfplotsset{
  log ticks with fixed point,
}
\def\addlegendimage{\csname pgfplots@addlegendimage\endcsname}

\title{Unintended Memorization in Large ASR Models, and How to Mitigate It}
\name{Lun Wang\quad\quad\quad Om Thakkar\quad\quad\quad Rajiv Mathews}
\address{Google}

\begin{document}

\maketitle

\begin{abstract}
It is well-known that neural networks can unintentionally memorize their training examples, causing privacy concerns. However, auditing memorization in large non-auto-regressive automatic speech recognition (ASR) models has been challenging due to the high compute cost of existing methods such as hardness calibration. In this work, we design a simple auditing method to measure memorization in large ASR models without the extra compute overhead. Concretely, we speed up randomly-generated utterances to create a mapping between vocal and text information that is difficult to learn from typical training examples. Hence, accurate predictions only for sped-up training examples can serve as clear evidence for memorization, and the corresponding accuracy can be used to measure memorization. Using the proposed method, we showcase memorization in the state-of-the-art ASR models. To mitigate memorization, we tried gradient clipping during training to bound the influence of any individual example on the final model. We empirically show that clipping each example’s gradient can mitigate memorization for sped-up training examples with up to 16 repetitions in the training set. Furthermore, we show that in large-scale distributed training, clipping the average gradient on each compute core maintains neutral model quality and compute cost while providing strong privacy protection.
\end{abstract}

\section{Introduction}
\label{sec:intro}

Neural networks can unintentionally memorize specific parts about their training examples.
Prior works demonstrated that auto-regressive language models~\cite{carlini2019secret,carlini2022quantifying} and vision models~\cite{carlini2023extracting} are susceptible to unintended memorization of their training examples, and thus may disclose potentially sensitive information during inference.
However, for non-auto-regressive models, memorization can be hard to distinguish from generalization.
For example, when a non-auto-regressive ASR model accurately transcribes a training example, it is hard to tell whether the model is generalizing well or it has unintentionally memorized the example.
The reason is that the difference in accuracy between the two cases is so small that it easily gets hidden by other sources of variance such as inherent hardness of different training examples (\emph{e.g.} some training examples are intrinsically easier/harder to learn than other examples and thus have higher/lower accuracy).
Existing works train ``reference'' models~\cite{jagielski2022measuring,tramer2022truth} to calibrate the hardness of different training examples to rule out the variance and bring out the subtle difference between memorization and generalization.
However, these works have to train tens to hundreds of reference models for obtaining good calibration.
As the size of trained models increases, obtaining comparable reference models can be very cost/compute/memory intensive.
Thus, a way to efficiently measure unintended memorization for large non-auto-regressive ASR models is urgently needed.

In this work, we propose the first efficient memorization auditing framework for large non-auto-regressive ASR models.
To address the ambiguity between generalization and memorization, we propose to create out-of-distribution training examples that are extremely hard to be learnt from normal training examples, such that the model can only memorize them for accurate transcription.
To obtain such training examples, we speed up normal utterances to create a mapping between vocal and text information different from typical training examples.
On the state-of-the-art ASR models~\cite{chiu2022self}, we manage to show that these training examples are unique enough such that memorization is the only way to accurately transcribe them.
As a result, accurate transcripts given by the ASR model for sped-up training examples can serve as clear evidence for memorization, and the level of accuracy can be used as a measure of memorization. 

To mitigate memorization, we propose to apply \emph{per-example gradient clipping} during training.
Specifically, we clip each training example's gradient to a fixed L2 norm bound if it's originally larger than the bound.
The intuition is that using per-example clipping, how much an individual example can influence the final model is bounded, and thus the final model should not memorize too much about any training example.
Our evaluation on the fine-tuning of the state-of-the-art BEST-RQ~\cite{chiu2022self} pre-trained ASR models shows that per-example clipping can effectively mitigate memorization for training examples occurring up to 16 times in the training set.
However, per-example clipping incurs extra training time overhead because we can no longer avoid materializing per-example gradients for acceleration like in non-private training.
To address the issue, we revisit the idea of \emph{micro-batch clipping}~\cite{ponomareva2023dp}, which shards the gradients into several micro-batches, averages the gradients within the same micro-batch, and then applies clipping.
Coincidentally, in large-scale distributed training, each compute core (\emph{e.g.} TPU/GPU) has to maintain the average gradient of all the training examples on it and thus forms natural micro-batches without incurring extra overhead.
Our empirical results show that \emph{per-core gradient clipping} achieves comparable or even better word error rate (WER) and neutral running time compared with the non-private baseline while providing much better empirical privacy.

In Section~\ref{sec:method}, we introduce our memorization auditing method and empirically evaluate it.
Next in Section~\ref{sec:countermeasure}, we introduce the countermeasure, gradient clipping, and use the proposed auditing method to demonstrate the effectiveness of the countermeasure.
We conclude this work in Section~\ref{sec:conclusion}.
\section{Auditing Memorization in ASR Models}
\label{sec:method}

In this section, we first introduce the background and give an overview of related works on memorization auditing.
Then, we highlight the unique challenge of auditing large non-auto-regressive ASR models.
Finally, we propose to audit memorization in ASR models efficiently using sped-up utterances, and empirically verify the effectiveness on the state-of-the-art BEST-RQ pre-trained ASR models~\cite{chiu2022self}.

\subsection{Background \& Related Works}

The Secret Sharer framework~\cite{carlini2019secret} has been widely used to measure unintended memorization of textual data in language models~\cite{carlini2019secret,carlini2022quantifying}, and detect such memorization even when they are fused with acoustic models for ASR~\cite{huang2022detecting}.
Specifically, Secret Sharer inserts hand-crafted training examples following a certain distribution, namely \emph{canaries}, into the training set.
To measure the level of memorization, Secret Sharer measures accuracy metrics, such as perplexities, of the canaries inserted in the training set and compare them with the metrics of examples drawn from the same distribution but unseen during training (\emph{i.e. holdout set}).
If the model performs significantly better on the inserted canaries than on the holdout set, then it is strong evidence that the model memorizes the inserted canaries, and the rank of the inserted canary's metric among the holdout set can serve as a measure of memorization.
This intuition is formalized in the following definition of \emph{exposure}, the metric we use to measure memorization.

\begin{definition}[Carlini \emph{et al.}~\cite{carlini2019secret}]
Given a canary $c$, a model $\mathcal{M}$, and examples in a holdout set $r_i$, the exposure of $c$ is
\begin{equation*}
\textbf{exposure}_\mathcal{M}(c, \{r_i\}) = \log_2|\{r_i\}| - \log_2\textbf{rank}_\mathcal{M}(c, \{r_i\}),
\end{equation*}
 where $|\{r_i\}|$ is the size of the holdout set, and $\textbf{rank}_\mathcal{M}(c, \{r_i\})$ is the rank of canary $c$ among $r_i$ in terms of a metric of interest, such as perplexity, or character error rate.
\end{definition}

Recent work has designed methods to demonstrate that model updates in ASR training can leak potentially sensitive attributes like speaker identity~\cite{dang2022method} of utterances used in computing the updates.
For measuring memorization of utterances used for training ASR models, there is only one work~\cite{jagielski2022measuring} that uses the Secret Sharer framework but requires additional similarly-trained ASR models as reference models for hardness calibration \cite{watson2021importance,carlini2022membership} as mentioned in Section~\ref{sec:intro}.
Training such reference models can be compute-intensive, as one particularly requires at least ten reference models for good calibration. 
For example, in Section~\ref{sec:challenge}, we show that without reference models, their method~\cite{jagielski2022measuring} highly underestimates the amount of memorization in the model.
Thus, there is no method that can accurately measure memorization in ASR models efficiently (\emph{i.e.}, without using reference models).

There are also works that focus on designing attacks for extracting training data, from auto-regressive models like LMs~\cite{carlini2022quantifying} and Diffusion models~\cite{carlini2023extracting}.
However, attacks for auto-regressive models are not directly applicable to non-auto-regressive ASR models, and existing extraction attacks on ASR models~\cite{amid2022extracting} have been shown to work only for memorization from commonly-occurring structured components in the training data.
There are also works that design  attacks on image models~\cite{balle2022reconstructing} for reconstructing images that might have been memorized during training. 
However, there is no work on reconstruction of audio data used for training ASR models.

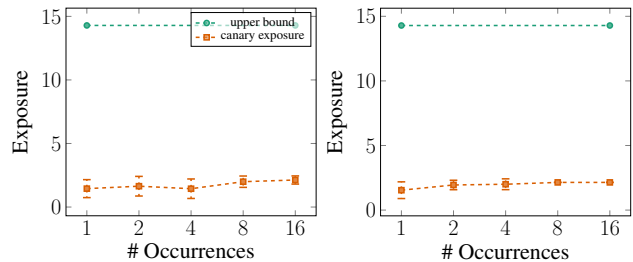
\begin{figure}[htb]
    \begin{subfigure}[b!]{0.23\textwidth}
    \begin{adjustbox}{width=\linewidth}
    \begin{tikzpicture}
    \begin{axis}[
        xmode=log,
        cycle list/Dark2,
        mark list fill={.!75!white},
        cycle multiindex* list={
            Dark2\nextlist
            my marks\nextlist
            dashed\nextlist
            very thick\nextlist
        },
        every axis plot/.append style={errorBars},
        samples=3,
        label style={font=\LARGE},
        xlabel={\# Occurrences},
        ylabel={Exposure},
        ticklabel style={
            font=\LARGE,
        },
        xtick={1, 2, 4, 8, 16},
        legend entries={upper bound, canary exposure},
        legend style={fill=white, fill opacity=0.6,text opacity=1},
    ]
        \addplot coordinates {(1, 14.2877123795) (16, 14.2877123795)};
        \addplot table[col sep=comma, x=occurrence, y=mean, y error=std] {data/exposure/bestrq300m/slow_seen.txt};

    \end{axis}
    \end{tikzpicture}
    \end{adjustbox}
    \label{fig:slow-300m}
    \caption{300M Conformer.}
    \end{subfigure}
    \begin{subfigure}[b!]{0.23\textwidth}
    \begin{adjustbox}{width=\linewidth}
    \begin{tikzpicture}
    \begin{axis}[
        xmode=log,
        cycle list/Dark2,
        mark list fill={.!75!white},
        cycle multiindex* list={
            Dark2\nextlist
            my marks\nextlist
            dashed\nextlist
            very thick\nextlist
        },
        every axis plot/.append style={errorBars},
        samples=3,
        label style={font=\LARGE},
        xlabel={\# Occurrences},
        ylabel={Exposure},
        ticklabel style={
            font=\LARGE,
        },
        xtick={1, 2, 4, 8, 16},
    ]
        \addplot coordinates {(1, 14.2877123795) (16, 14.2877123795)};
        \addplot table[col sep=comma, x=occurrence, y=mean, y error=std] {data/exposure/bestrq600m/slow_seen.txt};
    \end{axis}
    \end{tikzpicture}
    \end{adjustbox}
    \label{fig:slow-600m}
    \caption{600M Conformer.}
    \end{subfigure}

\caption{Exposure vs. \# occurrences for canaries from~\cite{jagielski2022measuring}. Upper bound is the exposure when the canary has better accuracy than all examples in the holdout set of size 20,000.}
\label{fig:slow}
\end{figure}

\subsection{Generalization or Memorization?}
\label{sec:challenge}

\noindent\textbf{Experiment Setup.} As mentioned in Section~\ref{sec:intro}, it can be challenging to distinguish between generalization and memorization for non-auto-regressive ASR models.
To demonstrate this, we conduct experiments by training state-of-the-art ASR models and audit memorization using the method in~\cite{carlini2019secret} without reference models.
Specifically, we choose the state-of-the-art ASR model architecture: the 600M Conformer XL~\cite{zhang2020pushing} and the 300M variant thereof.
The encoders are pre-trained on LibriLight~\cite{kahn2020libri} for 1M steps using self-supervised learning with random-projection quantizer (BEST-RQ)~\cite{chiu2022self}, and the complete model is fine-tuned on LibriSpeech~\cite{panayotov2015librispeech} for 20K steps.
Our training setup follows the parameter settings in the original BEST-RQ work.

We insert canaries in the fine-tuning phase to measure memorization.
Following the only prior work~\cite{jagielski2022measuring}, the transcript of the canaries is random combination of the top 10,000 words in LibriSpeech dataset.
Each transcript contains 7 words, the common length of LibriSpeech transcripts, and is inserted into the training set with  frequencies in $\{1, 2, 4, 8, 16\}$ to study the influence of repetition on memorization.
For each frequency, we create 20 unique transcripts to obtain error bars.
After getting the transcripts, the corresponding utterances are generated using a WaveNet Text-to-Speech (TTS) engine~\cite{oord2018parallel} using either male or female voice randomly.
We construct a holdout set composed of 20,000 examples following exactly the same recipe.

The metric used to calculate exposure is character-error-rate (CER), namely the editing distance between the ground truth and the transcript output by the model at the granularity of character.
We choose CER over word-error-rate (WER), the common metric for measuring ASR model quality, because it can capture character-level memorization.
For instance, if the model outputs ``Alis travols two Than Frensisko'' for an input utterance with ground truth ``Alice travels to San Francisco'', then CER can reflect some level of  memorization while WER will fail to recognize this as memorization.

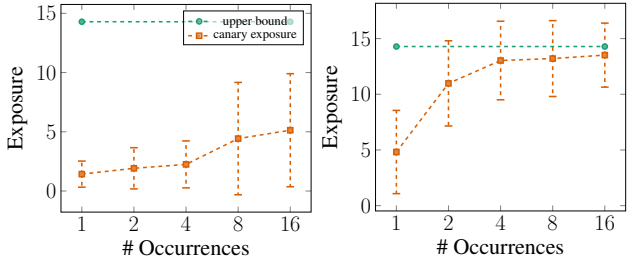
\begin{figure}
    \begin{subfigure}[b!]{0.23\textwidth}
    \begin{adjustbox}{width=\linewidth}
    \begin{tikzpicture}
    \begin{axis}[
        xmode=log,
        cycle list/Dark2,
        mark list fill={.!75!white},
        cycle multiindex* list={
            Dark2\nextlist
            my marks\nextlist
            dashed\nextlist
            very thick\nextlist
        },
        every axis plot/.append style={errorBars},
        samples=3,
        label style={font=\LARGE},
        xlabel={\# Occurrences},
        ylabel={Exposure},
        ticklabel style={
            font=\LARGE,
        },
        xtick={1, 2, 4, 8, 16},
        legend entries={upper bound, canary exposure},
        legend pos=north east,
        legend style={fill=white, fill opacity=0.6,text opacity=1},
    ]
        \addplot coordinates {(1, 14.2877123795) (16, 14.2877123795)};
        \addplot table[col sep=comma, x=occurrence, y=mean, y error=std] {data/exposure/bestrq300m/np.txt};
    \end{axis}
    \end{tikzpicture}
    \end{adjustbox}
    \label{fig:fast-300m}
    \caption{300M Conformer.}
    \end{subfigure}
    \begin{subfigure}[b!]{0.23\textwidth}
    \begin{adjustbox}{width=\linewidth}
    \begin{tikzpicture}
    \begin{axis}[
        xmode=log,
        cycle list/Dark2,
        mark list fill={.!75!white},
        cycle multiindex* list={
            Dark2\nextlist
            my marks\nextlist
            dashed\nextlist
            very thick\nextlist
        },
        every axis plot/.append style={errorBars},
        samples=3,
        label style={font=\LARGE},
        xlabel={\# Occurrences},
        ylabel={Exposure},
        ticklabel style={
            font=\LARGE,
        },
        xtick={1, 2, 4, 8, 16},
    ]
        \addplot coordinates {(1, 14.2877123795) (16, 14.2877123795)};
        \addplot table[col sep=comma, x=occurrence, y=mean, y error=std] {data/exposure/bestrq600m/np.txt};
    \end{axis}
    \end{tikzpicture}
    \end{adjustbox}
    \label{fig:fast-600m}
    \caption{600M Conformer.}
    \end{subfigure}

\caption{Exposure vs. \# occurrences for sped-up canaries.}
\label{fig:fast}
\end{figure}

\noindent\textbf{Evaluation Results.}
The evaluation results are summarized in Figure~\ref{fig:slow}, where we plot the exposure of the 300M and 600M models.
We observe that the models have low exposure when fine-tuned with canaries.
However, this does not mean the model does not memorize the canaries.
Instead, the reason is that even a model that never sees a canary can perform reasonably well in transcribing the canaries because the model generalizes well to learn how to transcribe the canaries used in~\cite{jagielski2022measuring} from normal training examples, and leaves no much room for improvement due to canary insertions for memorization measurement.
This ambiguity between generalization and memorization significantly reduces the power of the measurements for unintended memorization in large non-auto-regressive ASR models.

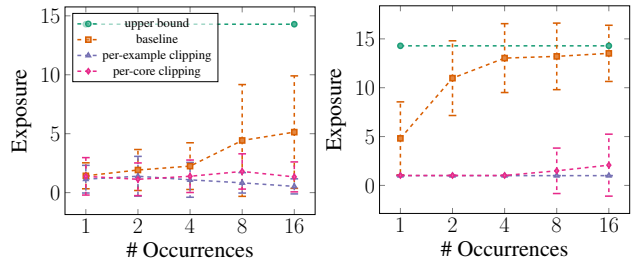
\begin{figure}[htbp]
    \begin{subfigure}[b!]{0.23\textwidth}
    \begin{adjustbox}{width=\linewidth}
    \begin{tikzpicture}
    \begin{axis}[
        xmode=log,
        cycle list/Dark2,
        mark list fill={.!75!white},
        cycle multiindex* list={
            Dark2\nextlist
            my marks\nextlist
            dashed\nextlist
            very thick\nextlist
        },
        every axis plot/.append style={errorBars},
        samples=3,
        label style={font=\LARGE},
        xlabel={\# Occurrences},
        ylabel={Exposure},
        ticklabel style={
            font=\LARGE,
        },
        xtick={1, 2, 4, 8, 16},
        legend entries={upper bound, baseline, per-example clipping, per-core clipping},
        legend pos=north west,
        legend style={fill=white, fill opacity=0.6,text opacity=1},
    ]
        \addplot coordinates {(1, 14.2877123795) (16, 14.2877123795)};
        \addplot table[col sep=comma, x=occurrence, y=mean, y error=std] {data/exposure/bestrq300m/np.txt};
        \addplot table[col sep=comma, x=occurrence, y=mean, y error=std] {data/exposure/bestrq300m/pec2p5.txt};
        \addplot table[col sep=comma, x=occurrence, y=mean, y error=std] {data/exposure/bestrq300m/pcc2p5.txt};
    \end{axis}
    \end{tikzpicture}
    \end{adjustbox}
    \label{fig:exposure-300m}
    \caption{300M Conformer.}
    \end{subfigure}
    \begin{subfigure}[b!]{0.23\textwidth}
    \begin{adjustbox}{width=\linewidth}
    \begin{tikzpicture}
    \begin{axis}[
        xmode=log,
        cycle list/Dark2,
        mark list fill={.!75!white},
        cycle multiindex* list={
            Dark2\nextlist
            my marks\nextlist
            dashed\nextlist
            very thick\nextlist
        },
        every axis plot/.append style={errorBars},
        samples=3,
        label style={font=\LARGE},
        xlabel={\# Occurrences},
        ylabel={Exposure},
        ticklabel style={
            font=\LARGE,
        },
        xtick={1, 2, 4, 8, 16},
    ]
        \addplot coordinates {(1, 14.2877123795) (16, 14.2877123795)};
        \addplot table[col sep=comma, x=occurrence, y=mean, y error=std] {data/exposure/bestrq600m/np.txt};
        \addplot table[col sep=comma, x=occurrence, y=mean, y error=std] {data/exposure/bestrq600m/pec2p5.txt};
        \addplot table[col sep=comma, x=occurrence, y=mean, y error=std] {data/exposure/bestrq600m/pcc2p5.txt};
    \end{axis}
    \end{tikzpicture}
    \end{adjustbox}
    \label{fig:exposure-600m}
    \caption{600M Conformer.}
    \end{subfigure}

\caption{Exposure vs. \# occurrences for per-example or per-core clipping.}
\label{fig:exposure}
\end{figure}

\begin{table*}[t]
    \centering
    \begin{tabular}{|c|c|c|c|}
        \hline
        Model Size & Training Method & WER at 20K & Smoothed steps/sec at 20K\\
        \hline
        \multirow{3}{*}{300M} & Non-private & \textbf{4.44} & 2.50 \\
        \cline{2-4}
         & Per-example clipping & 4.95 & 1.74 \\
        \cline{2-4}
         & Per-core clipping & 4.57 & \textbf{2.55} \\
        \hline
        \multirow{3}{*}{600M} & Non-private & 4.00 & \textbf{2.03} \\
        \cline{2-4}
         & Per-example clipping & 4.09 & 0.95\\
        \cline{2-4}
         & Per-core clipping & \textbf{3.87} & 1.96 \\
        \hline
    \end{tabular}
    \caption{Utility metrics for different training methods. Per-core batch size is 4 for both models.}
    \label{tab:utility_metrics}
\end{table*}
\subsection{Measuring Memorization using Fast Utterances}

To address the issue discussed above, we propose to create \emph{extremely fast} utterances as canaries.
Concretely, we follow the canary generation process described in Section~\ref{sec:challenge} except that we configure the TTS engine to greatly speed up utterances during generation.
The intuition is that the utterances are so fast that we hardly expect to encounter them in human speech and consequently ASR training data so the model can only memorize them for accurate transcription.
In this way, we manage to decouple memorization from generalization and efficiently measure unintended memorization in ASR models using the Secret Sharer framework~\cite{carlini2019secret}.

\noindent\textbf{Evaluation Results.}
To validate the effectiveness of the sped-up canaries, we insert 4x-sped-up canaries into the fine-tuning phase of the 300M and 600M Conformer models.
First, we observe that a model that never sees sped-up canaries during training performs poorly on them (\emph{e.g.} CER close to 1.0).
This confirms our conjecture that the models cannot generalize to sped-up canaries by learning from normal-paced training examples.
Second, we observe that models fine-tuned with sped-up canaries can accurately transcribe the canaries seen during training but cannot generalize to the examples in the sped-up holdout set, and thus exhibit high exposure as shown in Figure~\ref{fig:fast}.
Third, we observe that the more frequently a canary appears during training, the higher its exposure is.
For example, for 600M Conformer model, when a canary occurred 16 times in the training set, its exposure almost always reaches the upper bound, flagging severe unintended memorization.
In summary, using sped-up canaries, we managed to find clear evidence of unintended memorization and furthermore measure it without using any reference models.

\section{Towards Mitigation via Sensitivity-bounded Training}
\label{sec:countermeasure}

One established way to mitigate unintended memorization is to use differentially private (DP) training method~\cite{abadi2016deep}.
However, DP training suffers from the curse of dimensionality and tend to have unacceptably low utility on large ASR models~\cite{ganesh2023public}.
To achieve a balance between utility and privacy, we propose skip the noise addition and only keep the gradient clipping operation in DP training following~\cite{huang2022detecting}.
We show that while clipping each example's gradient shows stronger robustness to memorization, clipping the average gradient of the examples on the same compute core (\emph{e.g.} GPU/TPU) can achieve good robustness to memorization, neutral or even better WER, and neutral computation cost compared to the SotA non-private baseline at the same time under the large-scale distributed training scenario.

\subsection{Bounding Influence from Individual Examples}
One (but not the only) gradient clipping method that has been shown to be capable of mitigating unintended memorization is per-example clipping~\cite{huang2022detecting}, where the gradient of each example in a mini-batch is clipped before being averaged.
The intuition is that with per-example clipping, any training example is prevented from having an out-sized impact on the training procedure, consequently limiting the impact on the final trained model. 
Per-example clipping also provides a distinct advantage over ad-hoc methods for privacy protection against memorization, in that they provide a clear path towards achieving differential privacy guarantees. 
Concretely, by adding noise to the gradients after per-example clipping, we can rigorously prove that the model satisfies DP and is robust to memorization if necessary.

However, per-example clipping is known to slow down training~\cite{ponomareva2023dp} because it requires materializing per-example gradients, which is usually not necessary in the non-private counterpart.
Per-example clipping might also hurt the model utility because clipping changes the direction of the average gradient.

\noindent\textbf{Evaluation Results.}
As shown in Figure~\ref{fig:exposure}, after applying per-example clipping, the exposure of all the sped-up canaries with different frequencies is suppressed to a low level (\emph{i.e.} around 1.0) for both the 300M and 600M models.
This means the previously observed memorization is almost completely removed after adding per-example clipping.
On the other hand, from Table~\ref{tab:utility_metrics}, we observe that the excellent robustness to memorization comes with a cost in model accuracy and training time.
For the 300M model, the WER relatively increases by 11.5\% while the running time slows down by 30.4\%.
For the 600M model, the WER relatively increases by 2.3\% while the running time slows down by 53.2\%.

\vspace{-5pt}
\subsection{Towards Neutral WER and Running Time}

To narrow the gap in WER and running, we propose a relaxed version of per-example clipping, namely per-core clipping. 
Per-core clipping is a special instantiation of micro-batch clipping~\cite{ponomareva2023dp} under the scenario of large-scale distributed training.
Instead of clipping each example's gradient, per-core clipping clips the average gradient on each compute core (\emph{e.g.} GPU/TPU core).
As each compute core would have maintained the per-core average gradient, per-core clipping only adds the negligible cost of the clipping operation.

\noindent\textbf{Evaluation Results.}
As shown in Figure~\ref{fig:exposure}, per-core clipping shows robustness to memorization close to per-example clipping.
Although for a higher number of occurrences, per-core clipping is less robust than per-example clipping, it shows excellent performance in terms of WER and running time.
As shown in Table~\ref{tab:utility_metrics}, per-core clipping always matches the running time of the non-private baseline.
In terms of utility, per-core clipping is significantly better than per-example clipping.
For the 600M model, it even surpasses the non-private baseline with WER of 3.87.

\noindent\textbf{Remark on per-core batch size.}
Note that the memorization robustness of per-core clipping decreases as per-core batch size increases.
For both the 300M and 600M models, we use the default per-core batch size 4.
\section{Conclusion}
\label{sec:conclusion}

In this work, we designed the first efficient memorization auditing framework for large ASR models using fast utterances.
We conduct experiments on the-state-of-the-art ASR models and successfully measure memorization in these models.
As mitigation, we show that per-example clipping and per-core clipping effectively eliminate the previously shown memorization under different scenarios.

\newpage
\bibliographystyle{IEEEbib}
\bibliography{refs}

\begin{thebibliography}{10}

\bibitem{carlini2019secret}
Nicholas Carlini, Chang Liu, {\'U}lfar Erlingsson, Jernej Kos, and Dawn Song,
\newblock ``The secret sharer: Evaluating and testing unintended memorization
  in neural networks.,''
\newblock in {\em USENIX Security Symposium}, 2019.

\bibitem{carlini2022quantifying}
Nicholas Carlini, Daphne Ippolito, Matthew Jagielski, Katherine Lee, Florian
  Tramer, and Chiyuan Zhang,
\newblock ``Quantifying memorization across neural language models,''
\newblock {\em arXiv preprint arXiv:2202.07646}, 2022.

\bibitem{carlini2023extracting}
Nicholas Carlini, Jamie Hayes, Milad Nasr, Matthew Jagielski, Vikash Sehwag,
  Florian Tramer, Borja Balle, Daphne Ippolito, and Eric Wallace,
\newblock ``Extracting training data from diffusion models,''
\newblock {\em arXiv preprint arXiv:2301.13188}, 2023.

\bibitem{jagielski2022measuring}
Matthew Jagielski, Om~Thakkar, Florian Tramer, Daphne Ippolito, Katherine Lee,
  Nicholas Carlini, Eric Wallace, Shuang Song, Abhradeep Thakurta, Nicolas
  Papernot, et~al.,
\newblock ``Measuring forgetting of memorized training examples,''
\newblock {\em International Conference on Machine Learning}, 2022.

\bibitem{tramer2022truth}
Florian Tram{\`e}r, Reza Shokri, Ayrton San~Joaquin, Hoang Le, Matthew
  Jagielski, Sanghyun Hong, and Nicholas Carlini,
\newblock ``Truth serum: Poisoning machine learning models to reveal their
  secrets,''
\newblock in {\em Proceedings of the 2022 ACM SIGSAC Conference on Computer and
  Communications Security}, 2022, pp. 2779--2792.

\bibitem{chiu2022self}
Chung-Cheng Chiu, James Qin, Yu~Zhang, Jiahui Yu, and Yonghui Wu,
\newblock ``Self-supervised learning with random-projection quantizer for
  speech recognition,''
\newblock in {\em International Conference on Machine Learning}, 2022.

\bibitem{ponomareva2023dp}
Natalia Ponomareva, Hussein Hazimeh, Alex Kurakin, Zheng Xu, Carson Denison,
  H~Brendan McMahan, Sergei Vassilvitskii, Steve Chien, and Abhradeep~Guha
  Thakurta,
\newblock ``How to dp-fy ml: A practical guide to machine learning with
  differential privacy,''
\newblock {\em Journal of Artificial Intelligence Research}, vol. 77, 2023.

\bibitem{huang2022detecting}
W~Ronny Huang, Steve Chien, Om~Thakkar, and Rajiv Mathews,
\newblock ``Detecting unintended memorization in language-model-fused asr,''
\newblock {\em Interspeech}, 2022.

\bibitem{dang2022method}
Trung Dang, Om~Thakkar, Swaroop Ramaswamy, Rajiv Mathews, Peter Chin, and
  Fran{\c{c}}oise Beaufays,
\newblock ``A method to reveal speaker identity in distributed asr training,
  and how to counter it,''
\newblock in {\em ICASSP 2022-2022 IEEE International Conference on Acoustics,
  Speech and Signal Processing (ICASSP)}. IEEE, 2022, pp. 4338--4342.

\bibitem{watson2021importance}
Lauren Watson, Chuan Guo, Graham Cormode, and Alex Sablayrolles,
\newblock ``On the importance of difficulty calibration in membership inference
  attacks,''
\newblock {\em arXiv preprint arXiv:2111.08440}, 2021.

\bibitem{carlini2022membership}
Nicholas Carlini, Steve Chien, Milad Nasr, Shuang Song, Andreas Terzis, and
  Florian Tramer,
\newblock ``Membership inference attacks from first principles,''
\newblock in {\em 2022 IEEE Symposium on Security and Privacy (SP)}. IEEE,
  2022, pp. 1897--1914.

\bibitem{amid2022extracting}
Ehsan Amid, Om~Thakkar, Arun Narayanan, Rajiv Mathews, and Fran{\c{c}}oise
  Beaufays,
\newblock ``Extracting targeted training data from asr models, and how to
  mitigate it,''
\newblock {\em Interspeech}, 2022.

\bibitem{balle2022reconstructing}
Borja Balle, Giovanni Cherubin, and Jamie Hayes,
\newblock ``Reconstructing training data with informed adversaries,''
\newblock in {\em 2022 IEEE Symposium on Security and Privacy (SP)}. IEEE,
  2022, pp. 1138--1156.

\bibitem{zhang2020pushing}
Yu~Zhang, James Qin, Daniel~S Park, Wei Han, Chung-Cheng Chiu, Ruoming Pang,
  Quoc~V Le, and Yonghui Wu,
\newblock ``Pushing the limits of semi-supervised learning for automatic speech
  recognition,''
\newblock {\em arXiv preprint arXiv:2010.10504}, 2020.

\bibitem{kahn2020libri}
Jacob Kahn, Morgane Rivi{\`e}re, Weiyi Zheng, Evgeny Kharitonov, Qiantong Xu,
  Pierre-Emmanuel Mazar{\'e}, Julien Karadayi, Vitaliy Liptchinsky, Ronan
  Collobert, Christian Fuegen, et~al.,
\newblock ``Libri-light: A benchmark for asr with limited or no supervision,''
\newblock in {\em ICASSP 2020-2020 IEEE International Conference on Acoustics,
  Speech and Signal Processing (ICASSP)}, 2020.

\bibitem{panayotov2015librispeech}
Vassil Panayotov, Guoguo Chen, Daniel Povey, and Sanjeev Khudanpur,
\newblock ``Librispeech: an asr corpus based on public domain audio books,''
\newblock in {\em 2015 IEEE international conference on acoustics, speech and
  signal processing (ICASSP)}. IEEE, 2015, pp. 5206--5210.

\bibitem{oord2018parallel}
Aaron Oord, Yazhe Li, Igor Babuschkin, Karen Simonyan, Oriol Vinyals, Koray
  Kavukcuoglu, George Driessche, Edward Lockhart, Luis Cobo, Florian Stimberg,
  et~al.,
\newblock ``Parallel wavenet: Fast high-fidelity speech synthesis,''
\newblock in {\em International conference on machine learning}. PMLR, 2018,
  pp. 3918--3926.

\bibitem{abadi2016deep}
Martin Abadi, Andy Chu, Ian Goodfellow, H~Brendan McMahan, Ilya Mironov, Kunal
  Talwar, and Li~Zhang,
\newblock ``Deep learning with differential privacy,''
\newblock in {\em Proceedings of the 2016 ACM SIGSAC conference on computer and
  communications security}, 2016, pp. 308--318.

\bibitem{ganesh2023public}
Arun Ganesh, Mahdi Haghifam, Milad Nasr, Sewoong Oh, Thomas Steinke,
  Om~Thakkar, Abhradeep~Guha Thakurta, and Lun Wang,
\newblock ``Why is public pretraining necessary for private model training?,''
\newblock in {\em International Conference on Machine Learning}. PMLR, 2023,
  pp. 10611--10627.

\end{thebibliography}

\end{document}